%% file: main.tex
\documentclass{article}
\usepackage{spconf,amsmath,graphicx}
\usepackage{amssymb}
\usepackage{subfig}
\usepackage[hidelinks]{hyperref}
\usepackage{color}
\usepackage{changes,siunitx}

\title{MANet: Improving Video Denoising with a Multi-Alignment Network}
\name{ Yaping Zhao$^{1,3,*}$
 , Haitian Zheng$^{2,*}$, Zhongrui Wang$^{1,3}$, Jiebo Luo$^{2}$, Edmund Y. Lam$^{1,3,\dag}$}

\address{
$^1$ The University of Hong Kong \, \indent \,
$^2$ University of Rochester\\
$^3$ACCESS –- AI Chip Center for Emerging Smart Systems}

\begin{document}
\input{math_command}

\twocolumn[{%
\renewcommand\twocolumn[1][]{#1}%
\maketitle
\begin{center}
\vspace{-15pt}
    \includegraphics[width=1.\linewidth]{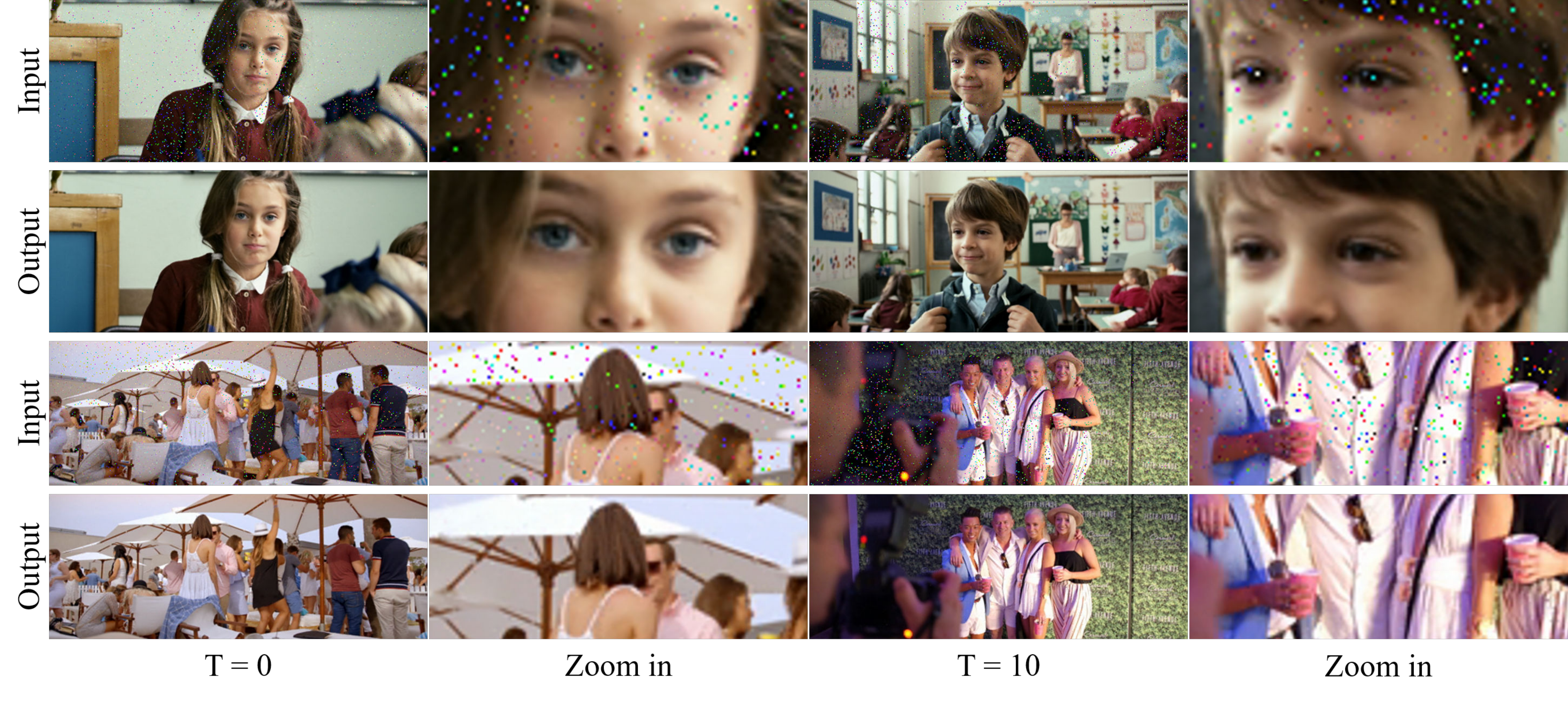}
\vspace{-20pt}
	\captionof{figure}{
	Given a noisy video sequence as input,
we propose a neural network using multiple alignments for video denoising to synthesize clean frames. 
	}
	\label{fig:teaser}
\end{center}
}
]

\newcommand\blfootnote[1]{%
  \begingroup
  \renewcommand\thefootnote{}\footnote{#1}%
  \addtocounter{footnote}{-1}%
  \endgroup
}

\blfootnote{$^{*}$Equal contribution.  $^{\dag}$Corresponding author.}
\blfootnote{This work is supported in part by ACCESS –- AI Chip Center for Emerging Smart Systems, Hong Kong SAR, in part by Hong Kong Research Grant Council (Grant No. 27206321), National Natural Science Foundation of China  (Grant No. 62122004).}

\begin{abstract}

In video denoising, the adjacent frames often provide very useful information, but accurate alignment is needed before such information can be harnassed. In this work, we present a multi-alignment network, which generates multiple flow proposals followed by attention-based averaging. It serves to mimic the non-local mechanism, suppressing noise by averaging multiple observations. Our approach can be applied to various state-of-the-art models that are based on flow estimation.
Experiments on a large-scale video dataset demonstrate that our method improves the denoising baseline model by \SI{0.2}{\decibel}, and further reduces the parameters by 47\% with model distillation.
Code is available at \href{https://github.com/IndigoPurple/MANet}{\textcolor{blue}{https://github.com/IndigoPurple/MANet}}.

\end{abstract}

\begin{keywords}
video denoising, image alignment, video enhancement, attention, image synthesis
\end{keywords}

\section{introduction}
\label{sec:intro}

\begin{figure*}
\centering
\vspace{-20pt}
    \includegraphics[width =0.81\linewidth]{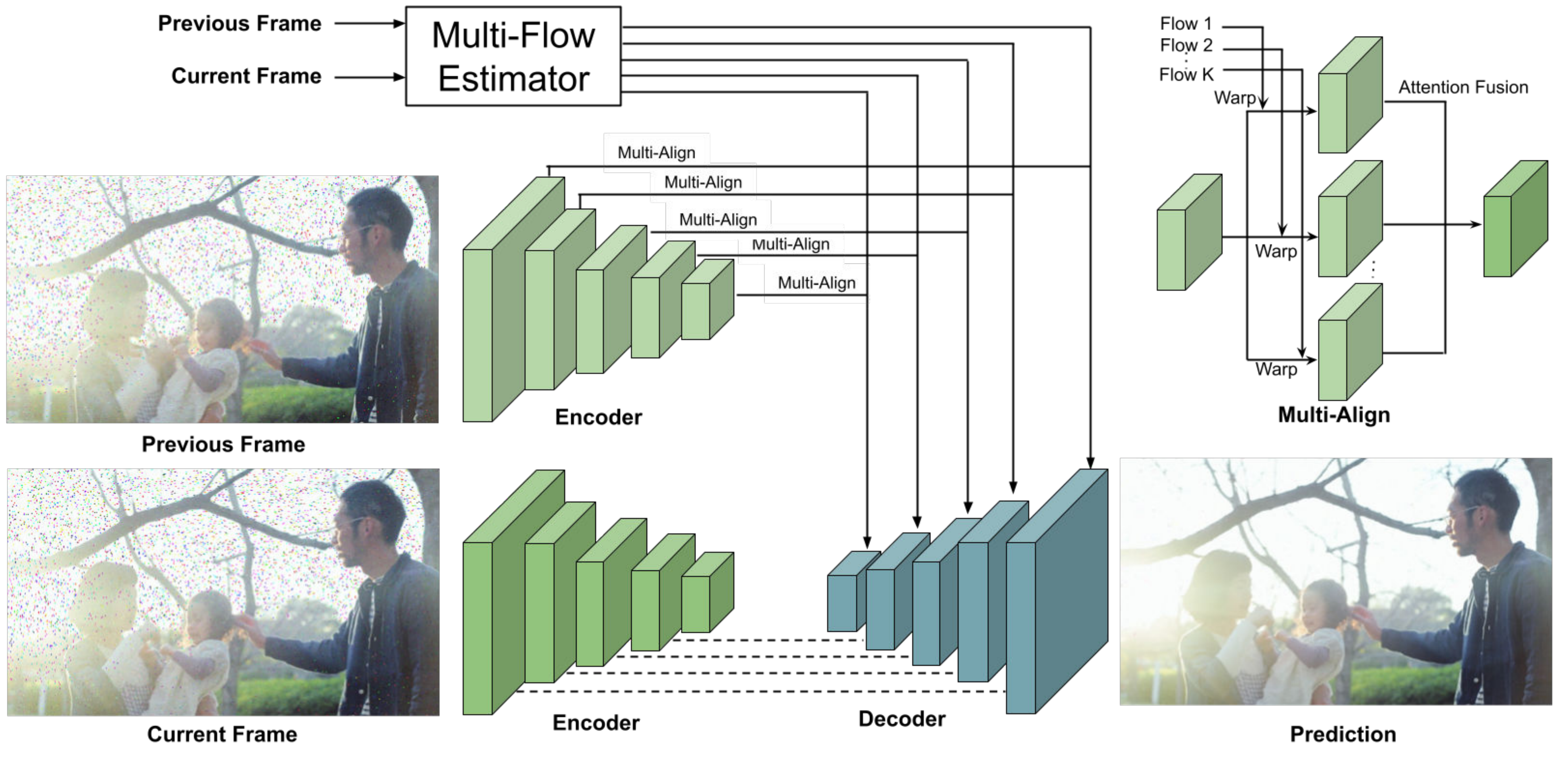}
    \vspace{-8pt}
    \caption{Network structure of the video denoising framework with multiple alignment.}
    \label{fig:method}
\end{figure*}

Video denoising aims to restore a clean video sequence from one that is corrupted with noise. 
Due to thermal effects, sensor imperfections or low-light, noise inevitably degrades the quality of many captured videos, for which video denoising becomes necessary. Additionally, it is also an indispensable sub-task of the remastering of vintage films, which are corrupted with a lot of noise because of technical limitations.

Unlike image denoising, video denoising can take advantage of information from adjacent frames. However, correspondence matching and frame alignment are crucial problems. Recent works~\cite{crossnet, xue2019video} rely on optical flow together with backward warping to perform such frame alignment. However, the existence of occlusion, motion blur, rotation or lighting change across different frames decreases the accuracy of many alignment methods.
Similarly, non-local means~\cite{wang2018non} is a powerful technique in image denoising because of its ability to suppress noise through averaging multiple observations of similar patches. In the context of video denoising, this approach would again rely on accurate frame alignment~\cite{crossnet, xue2019video, efenet}.

In this paper, we present a new video denoising method as Figure~\ref{fig:teaser} shows, together with the multiple alignments. To address the issue of information loss from misalignment, we propose to generate multiple optical flow candidates using a learning-based flow estimator. To mimic the non-local behavior for noise cancellation, we perform an attention-based averaging on the aligned feature.
Based on the designed architecture, we further perform model distillation to reduce the model size.
The resulting model is lightweight, and can be easily incorporated with recent methods.

To demonstrate the effectiveness of our approach, we perform experiments on a large-scale video dataset Vimeo~\cite{xue2019video}. Compared with the recent baseline~\cite{crossnet}, our approach improves the performance by \SI{0.2}{\decibel}, suggesting a possible direction to improve video denoising.

\section{Related Work}
\label{sec:related}

Let $x_{i}$ be the clean frame from a video sequence at timestep $T = i$, and $y_i = c(x_i)$ be an observed video frame through a
corrupted channel $c(\cdot)$. The task of video denoising is to utilize the addition information from the observed adjacent frames $Y_i(t) = \{y_j | j = i-t,\dots,i-1\}$ to restore a clean frame $\hat{x}_i$. Therefore, video denoising can be formalized by 
\begin{align}
    \hat{x}_i = \mathcal{D}(y_i, Y_i) .
\end{align}

Traditionally, the image restoration task can be formulated by the \emph{maximum a posteriori} (MAP) estimation under a predefined image prior. Typical image priors include hyper-Laplacian~\cite{krishnan2009fast}, mixture of Gaussian~\cite{zoran2011learning}, total variation~\cite{babacan2008total}, local linear embedding~\cite{chang2004super}, and sparsity~\cite{yang2008image}. With the availability of image restoration datasets, the supervised learning approach has also been explored. The typical approaches include decision tree~\cite{salvador2015naive}, random forest~\cite{schulter2015fast}, and dictionary learning~\cite{yang2013fast}. Recently, deep learning approaches~\cite{dong2014learning,kim2016accurate} further push the limit of the image restoration performance.

To utilize the adjacent frame information, video restoration approaches make use of the correspondence estimation to align them with the current frame. One approach is to perform patch motion compensation~\cite{tao2017detail}, while some others~\cite{crossnet, xue2019video} rely on optical flow. Specifically, let $\fv_{j\to i}$ be the estimated flow field from frame $j$ to $i$, Xue \textit{et al.}~\cite{xue2019video} performs backward warping on the adjacent frame, while Zheng \textit{et al.}~\cite{crossnet} additionally performs backward warping on the feature pyramid.

\section{Method}
\label{sec:method}

\subsection{Network Structure}
To perform multiple alignments and frame synthesis for video denoising, we employ the network structure of~\cite{crossnet} and replace its flow estimator and alignment module with our multi-alignment modules. The resulting network consists of two feature pyramid extractors, a multi-flow estimator, a multi-scale feature domain alignment module, and a UNet~\cite{unet} structure for frame synthesis, as
Figure~\ref{fig:method} depicts.

\subsection{Multiple Alignment}
For simplicity, we consider denoising the current frame using observation from the current  and the previous frame, \textit{i.e.},
\begin{align}
    \hat{x}_i = \mathcal{D}(y_i, y_{i-1}).
\end{align}
Nevertheless, it should be pointed out that our model can be easily extended to include multiple adjacent frames.

\noindent{\textbf{Multiple Flow Estimation. }}
Similar to~\cite{crossnet}, our approach relies on a learning-based flow estimator named FlowNet~\cite{flownet}. However, due to possible errors in the flow estimation, information from the adjacent frame is not efficiently used. Therefore, we use FlowNet to generate $K$ optical flows $\{ \fv^{(1)}_{i\to i-1},\dots,\fv^{(K)}_{i\to i-1} \}$ by increasing the channel of the flow output layer from $2$ to $2K$. By generating multiple possible flow estimations, the possibility of misalignment is reduced.

\noindent{\textbf{Image Alignment. }}
In the next step, we perform the spatial alignment $K$ times using backward warping. Specifically, we generate $K$ alignment with 
\begin{align}
\begin{aligned}
\label{eq:warp}
    &I^{(1)}_w = \mathrm{warp}(\fv^{(1)}_{i\to i-1}, y_{i-1}),\\
    &\vdots\\
    &I^{(K)}_w = \mathrm{warp}(\fv^{(K)}_{i\to i-1}, y_{i-1}),
\end{aligned}
\end{align}
where $\mathrm{warp}( \cdot, \cdot)$ denotes the warping operation implemented by the spatial transformer network~\cite{jaderberg2015spatial}.

\noindent{\textbf{Attention-based Averaging. }}
\label{sec:attention}
We further propose an attention-based averaging to rule out the misalignment. It serves to suppress noise by mimicking the non-local mechanism.

Specifically, we develop two different approaches to generate the unnormalized attention map, namely, \textbf{fc} and \textbf{ip} :

\begin{itemize}
    \item 
\textbf{fc} refers to fully-connected.
Specifically, we add an additional $1 \times 1$ convolutional layer with $K$ output channels to the flow estimator, which generates a $K$-channel unnormalized attention map $\{ b^{(1)},\dots,b^{(K)} \}$.

\item \textbf{ip} refers to inner-product. It is based on computing the similarities between the current frame $y_i$ and the warped previous frame $I_w^{(l)}$. Specifically, $y_i$ and $I_w^{(l)}$ are individually transferred to the feature maps $\theta(y_i)$ and $\phi(I_w^{(l)})$ by a $1 \times 1$ convolutional layer.
Then unnormalized attention maps are generated by inner product
\begin{align}
    b^{(l)} = \theta(y_i) ^\top \phi(I_w^{(l)}),\indent
    l = 1, \dots, K.
\end{align}
\end{itemize}

The normalized attention map is generated by performing the channel-wise softmax operation at each location $(m, n)$,
\textit{i.e.},
\begin{align}
    a_{mn}^{(l)} = \frac{e^{b_{mn}^{(l)}}}{e^{b_{mn}^{(1)}} + \dots + e^{b_{mn}^{(K)}}}, \indent
    l = 1, \dots, K.
\end{align}

The attention weights are then used for averaging the multiple aligned feature maps, where
\begin{align}
\label{eq:avg_align}
    I_a = \sum_{l=1}^K a^{(l)}\odot I^{(l)}_w.
\end{align}

It is worth noting that the alignment and attention fusion in Equations~\ref{eq:warp} and ~\ref{eq:avg_align} can be applied both on the image and the feature map, meaning that our multiple alignment framework can be incorporated into a wide range of models~\cite{crossnet, xue2019video, efenet}. 

\subsection{Model Distillation}
To reduce the model size, we cut the channel size of the flow estimator model and perform model distillation based on our trained model. We use the original loss to train the slimmed model. Additionally, we leverage an $\ell_1$ loss to minimize the feature and flow prediction difference between the two models to ensure that the slimmed model has a similar performance as the trained large model.

\section{Experiments}
\label{sec:exp}

\begin{figure}
    \includegraphics[width =\linewidth]{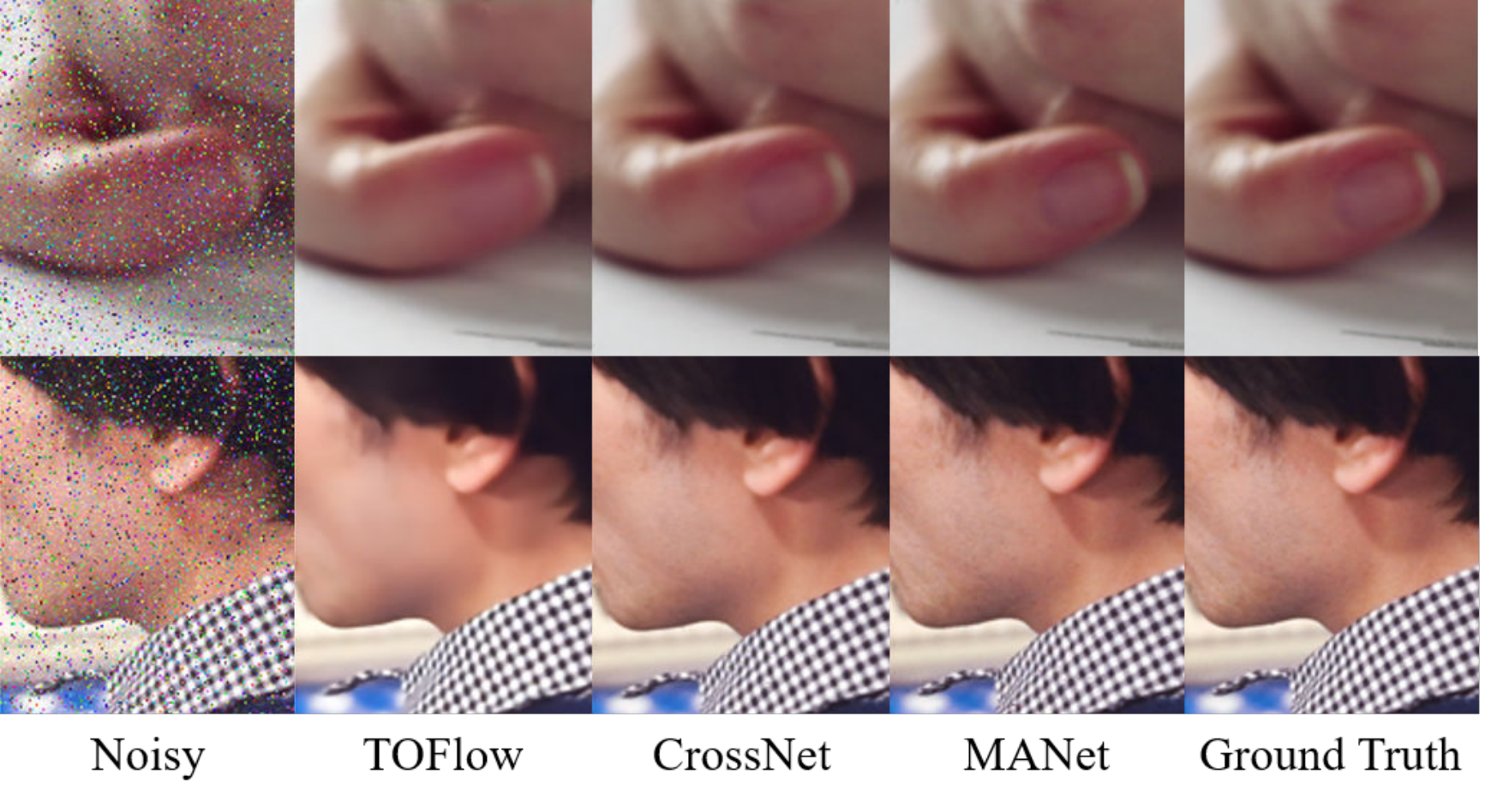}
    \vspace{-18pt}
    \caption{Video denoising comparisons between different algorithms on the Vimeo dataset. Zoom in to see details.}
    \label{fig:exp}
\end{figure}

\begin{table}
    \centering
    \resizebox{0.5\textwidth}{!}{
    \begin{tabular}{|c|c|c|c|c|c|}
    \hline
         Method &  TOFlow & CrossNet & MANet-fc & MANet-ip & MANet-ip-s\\
         \hline
       PSNR  & 33.51 & 45.02 & 45.11 & 45.23 & 45.22 \\
       \hline
       Parameters & 17.00 M & 35.18 M & 35.46 M & 35.45 M & 18.67 M\\
       \hline
    \end{tabular}
    }
    \caption{ Comparisons of denoising performance and parameter sizes among different models.}
    \label{tab:psnr}
\end{table}

\subsection{Denoising Performance Comparisons}
To validate the effectiveness of our  approach, we take CrossNet~\cite{crossnet} as the baseline and replace its alignment module with our proposed multi-alignment modules, resulting in \textit{MANet-fc} and \textit{MANet-ip}, as mentioned in Section~\ref{sec:attention}. Training and testing are performed on the Vimeo dataset~\cite{xue2019video} with mixed noise including a 10\% salt-and-pepper noise in addition to the Gaussian noise with a standard deviation (std) of 0.1, while the performance is measured by the PSNR metric. For intuitive qualitative comparisons, we additionally take the method TOFlow~\cite{xue2019video}, as Figure~\ref{fig:exp} shows.

As Table~\ref{tab:psnr} shows, the \textit{MANet-fc} improves the baseline by \SI{0.09}{\decibel} and the \textit{MANet-ip} improves the baseline by \SI{0.21}{\decibel}, while the slimmed model \textit{MANet-ip-s} after model distillation improves the baseline by \SI{0.2}{\decibel}.


\begin{figure*}
    \includegraphics[width =\linewidth]{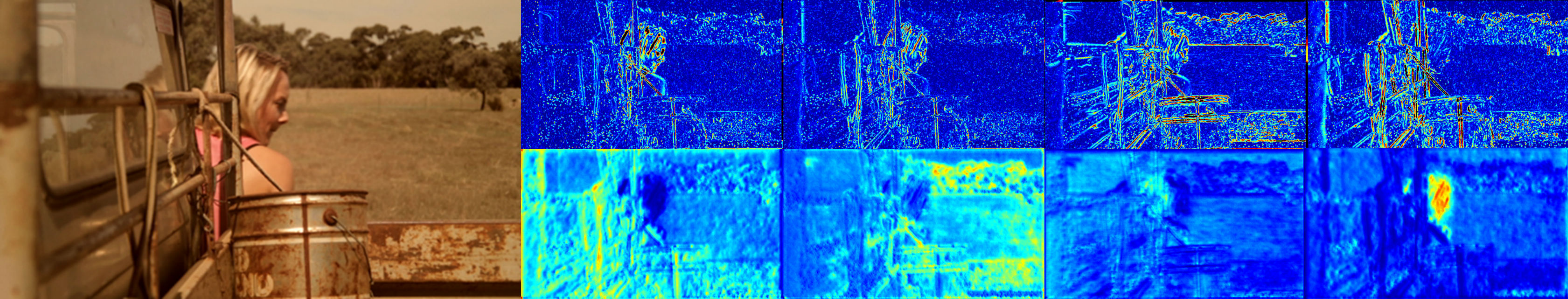}
    \caption{ The visualization of the ground truth (the first column), warping error (the first row on the right) and the corresponding attentional weight (the second row on the right).}
    \label{fig:vis_attention}
\end{figure*}

\begin{figure}
\centering
\vspace{-5pt}
    \includegraphics[width =0.8\linewidth]{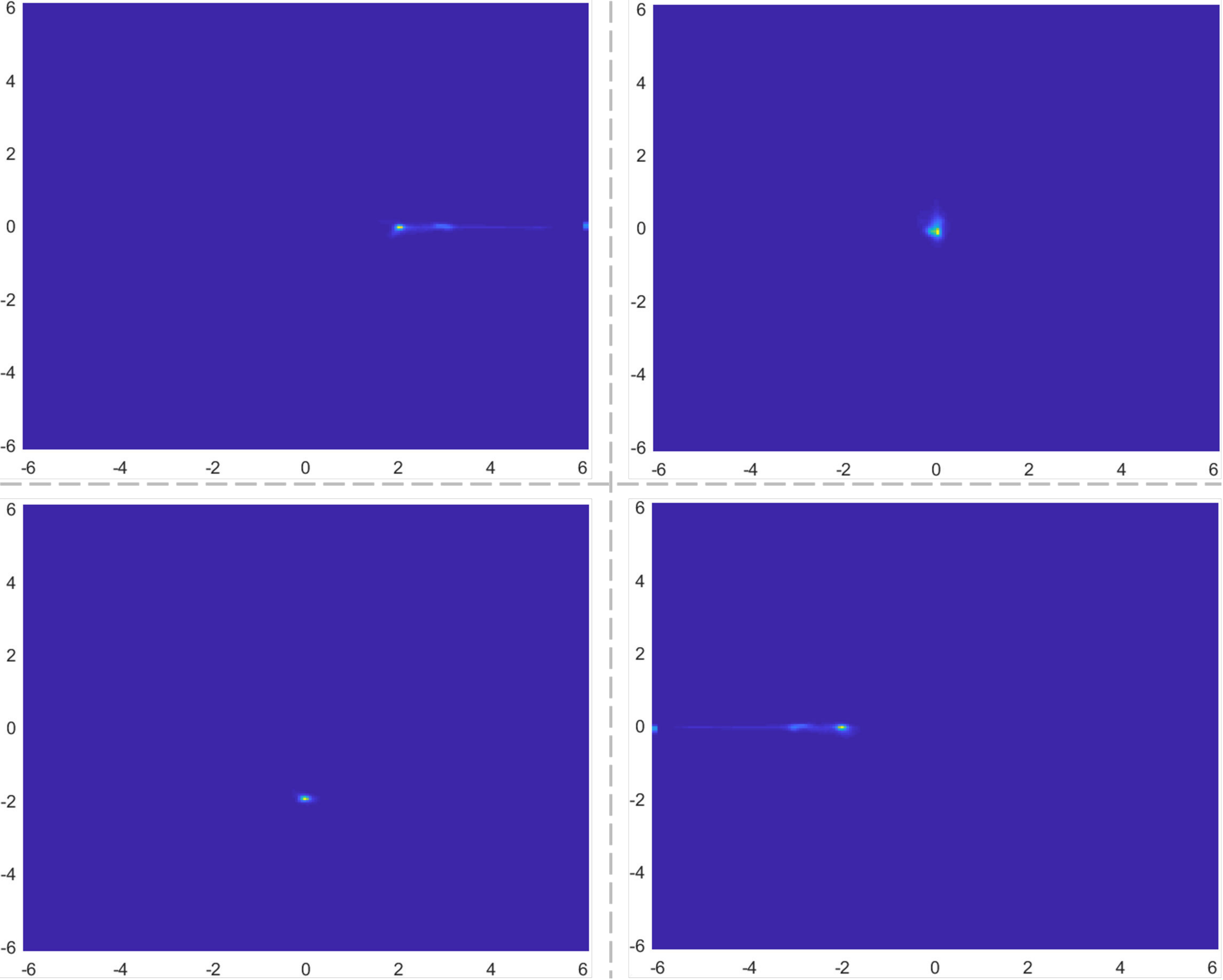}
    \caption{The visualization of the offset from multiple flows generated by MANet to CrossNet-generated flow.}
    \label{fig:offset}
\end{figure}

\subsection{Training Details}

All the models are trained from scratch for $11$ epochs using the Adam~\cite{kingma2014adam} optimizer. The learning rates are set to $3 \times 10^{-5}$ and decay by a factor of $0.1$ after every $4$ epochs.
In our experiment, the multiple alignment factor $K$ is set to $4$. We experimented other $K$ values and found that PSNR saturates after $K>4$, \textit{e.g.}, $K=8$ increases PSNR by $0.03$. To reduce memory usage, therefore, we set $K$ to the optimal value $4$.

\subsection{Visualization}
To further understand what the multiple alignment model has learned, we visualize the generated attention individually, and compare it with the error residue map by taking the difference between the reference frame and the current frame. Each row from the right side of Figure~\ref{fig:vis_attention} depicts an attention map with its warping error residue. From the second and the third columns, we can observe that the attention map tries to avoid the center misalignment region. It can also be  observed that the attention is evenly split to the $K$ attention maps for the smooth region, \textit{e.g.}, the top left glass region.

To understand the behavior of the multiple optical flows, we additionally visualize a 2-D histogram of the offset from multiple flows generated by MANet to CrossNet-generated flow. From Figure~\ref{fig:offset}, we can see that the learned multiple flow map tends to concentrate on the fixed offset. It suggests that for learning more adaptive multiple flow combinations, a more powerful flow estimator design is required.

\begin{figure}
\vspace{-5pt}
    \includegraphics[width =\linewidth]{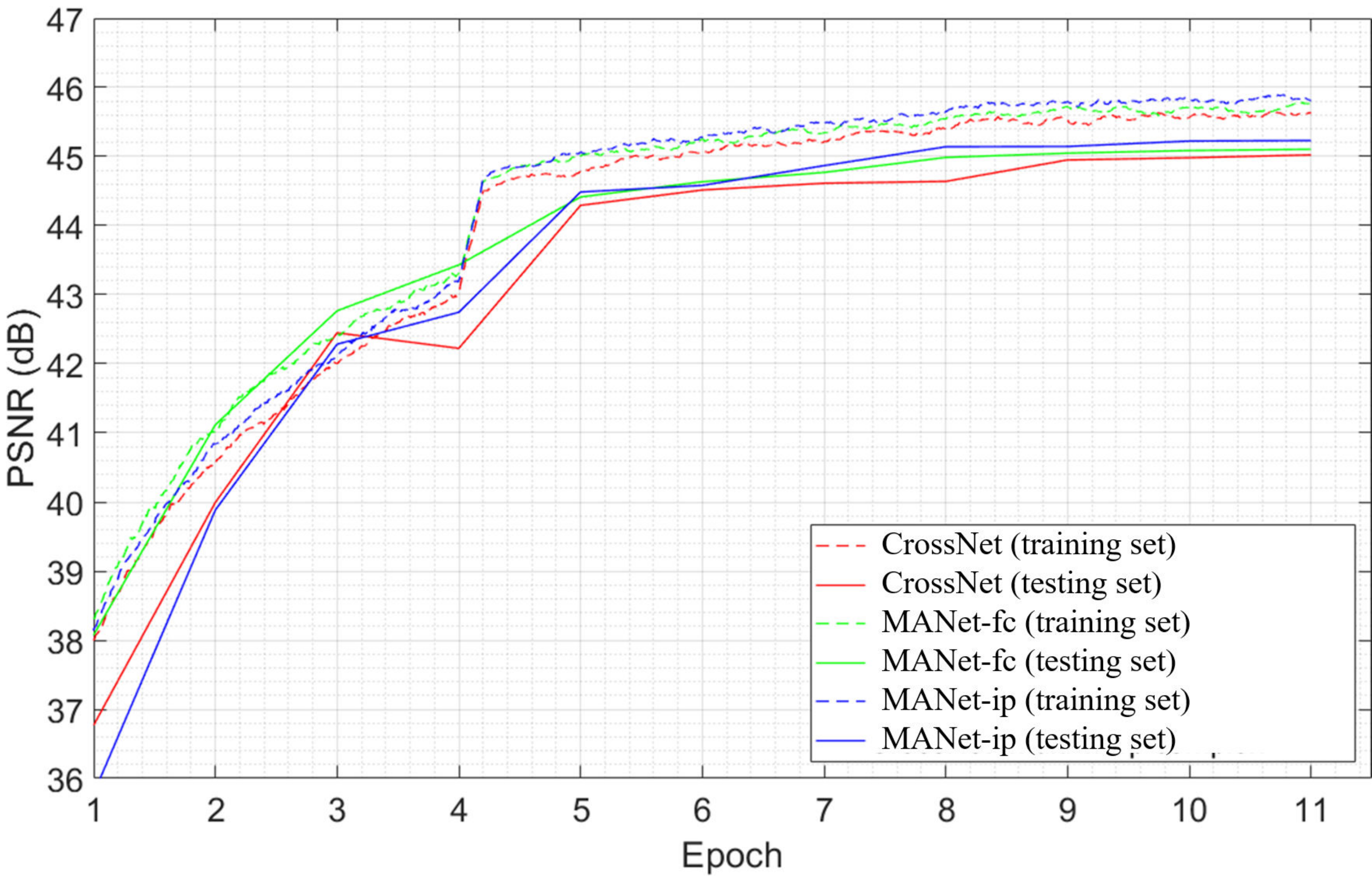}
    \caption{The convergence analysis on our proposed multiple alignment methods versus the baseline model. 
    }
    \label{fig:convergence}
\end{figure}

\subsection{Convergence}
As Figure~\ref{fig:convergence} shows, the \textit{MANet-fc} converges faster than the baseline mode consistently. The \textit{MANet-ip} converges slower than the other two models. However, it performs best at the end of the training. We speculate that the initial learning rate is too large for \textit{MANet-ip} to converge well.

\subsection{Network Efficiency}
Our model is based on generating $K$ flow estimations for enriched feature alignment. However, the additional parameters introduced are negligible as we only increase the flow prediction layer and the convolution layer from $2$ to $2K$, while adding a few more $1\times1$ convolutional layers for the aligned feature averaging in Eq.~\ref{eq:avg_align}.
The  size of the three models are shown in Table~\ref{tab:psnr}. Our models \textit{MANet-fc} and \textit{MANet-ip} only require $0.28M (0.79\%)$ and $0.27M (0.77\%)$ additional parameters in comparison to the baseline model. Furthermore, after model distillation, the slimmed model \textit{MANet-ip-s} reduces parameters by $47\%$.

\section{Conclusion}
In this paper, we present a multi-alignment network to address the misalignment issue and incorporate the non-local mean approach. The multiple flow estimation and alignment reduce the risk of misalignment. The attention-based averaging mimics the non-local component for effective denoising. Combined with knowledge distillation, model
parameters are reduced by 47\%. Experiments on a large-scale dataset show that our approach outperforms a recent baseline, suggesting a new angle for improving video denoising performance.

\bibliographystyle{IEEEbib}
\small{\bibliography{refs}}

\end{document}

%% file: math_command.tex
\newcommand{\Amat}{{\bf A}}
\newcommand{\Bmat}{{\bf B}}
\newcommand{\Cmat}{{\bf C}}
\newcommand{\Dmat}{{\bf D}}
\newcommand{\Emat}[0]{{{\bf E}}}
\newcommand{\Fmat}[0]{{{\bf F}}}
\newcommand{\Gmat}[0]{{{\bf G}}}
\newcommand{\Hmat}[0]{{{\bf H}}}
\newcommand{\Imat}{{\bf I}}
\newcommand{\Jmat}[0]{{{\bf J}}}
\newcommand{\Kmat}[0]{{{\bf K}}}
\newcommand{\Lmat}[0]{{{\bf L}}}
\newcommand{\Mmat}[0]{{{\bf M}}}
\newcommand{\Nmat}[0]{{{\bf N}}}
\newcommand{\Omat}[0]{{{\bf O}}}
\newcommand{\Pmat}[0]{{{\bf P}}}
\newcommand{\Qmat}[0]{{{\bf Q}}}
\newcommand{\Rmat}[0]{{{\bf R}}}
\newcommand{\Smat}[0]{{{\bf S}}}
\newcommand{\Tmat}[0]{{{\bf T}}}
\newcommand{\Umat}{{{\bf U}}}
\newcommand{\Vmat}[0]{{{\bf V}}}
\newcommand{\Wmat}[0]{{{\bf W}}}
\newcommand{\Xmat}{{\bf X}}
\newcommand{\Ymat}[0]{{{\bf Y}}}
\newcommand{\Zmat}{{\bf Z}}

\newcommand{\av}{\boldsymbol{a}}
\newcommand{\Av}{\boldsymbol{A}}
\newcommand{\Cv}{\boldsymbol{C}}
\newcommand{\bv}{\boldsymbol{b}}
\newcommand{\cv}{{\boldsymbol{c}}}
\newcommand{\dv}{\boldsymbol{d}}
\newcommand{\ev}[0]{{\boldsymbol{e}}}
\newcommand{\fv}{\boldsymbol{f}}
\newcommand{\Fv}[0]{{\boldsymbol{F}}}
\newcommand{\gv}[0]{{\boldsymbol{g}}}
\newcommand{\hv}[0]{{\boldsymbol{h}}}
\newcommand{\iv}[0]{{\boldsymbol{i}}}
\newcommand{\jv}[0]{{\boldsymbol{j}}}
\newcommand{\kv}[0]{{\boldsymbol{k}}}
\newcommand{\lv}[0]{{\boldsymbol{l}}}
\newcommand{\mv}[0]{{\boldsymbol{m}}}
\newcommand{\nv}{\boldsymbol{n}}
\newcommand{\ov}[0]{{\boldsymbol{o}}}
\newcommand{\pv}[0]{{\boldsymbol{p}}}
\newcommand{\qv}[0]{{\boldsymbol{q}}}
\newcommand{\rv}[0]{{\boldsymbol{r}}}
\newcommand{\sv}[0]{{\boldsymbol{s}}}
\newcommand{\tv}[0]{{\boldsymbol{t}}}
\newcommand{\uv}[0]{{\boldsymbol{u}}}
\newcommand{\vv}{\boldsymbol{v}}
\newcommand{\wv}{\boldsymbol{w}}
\newcommand{\Wv}{\boldsymbol{W}}
\newcommand{\xv}{\boldsymbol{x}}
\newcommand{\yv}{\boldsymbol{y}}
\newcommand{\Xv}{\boldsymbol{X}}
\newcommand{\Yv}{\boldsymbol{Y}}
\newcommand{\zv}{\boldsymbol{z}}

\newcommand{\Gammamat}[0]{{\boldsymbol{\Gamma}}}
\newcommand{\Deltamat}[0]{{\boldsymbol{\Delta}}}
\newcommand{\Thetamat}{\boldsymbol{\Theta}}
\newcommand{\Lambdamat}{{\boldsymbol{\Lambda}}}
\newcommand{\Ximat}[0]{{\boldsymbol{\Xi}}}
\newcommand{\Pimat}[0]{{\boldsymbol{\Pi}} }
\newcommand{\Sigmamat}{\boldsymbol{\Sigma}}
\newcommand{\Upsilonmat}[0]{{\boldsymbol{\Upsilon}} }
\newcommand{\Phimat}{\boldsymbol{\Phi}}
\newcommand{\Psimat}{\boldsymbol{\Psi}}
\newcommand{\Omegamat}{{\boldsymbol{\Omega}}}

\newcommand{\Lambdav}{\bm{\Lambda}}
\newcommand{\alphav}{\boldsymbol{\alpha}}
\newcommand{\betav}[0]{{\boldsymbol{\beta}} }
\newcommand{\gammav}{{\boldsymbol{\gamma}}}
\newcommand{\deltav}[0]{{\boldsymbol{\delta}} }
\newcommand{\epsilonv}{\boldsymbol{\epsilon}}
\newcommand{\zetav}[0]{{\boldsymbol{\zeta}} }
\newcommand{\etav}[0]{{\boldsymbol{\eta}} }
\newcommand{\thetav}{\boldsymbol{\theta}}
\newcommand{\iotav}[0]{{\boldsymbol{\iota}} }
\newcommand{\kappav}{{\boldsymbol{\kappa}}}
\newcommand{\lambdav}[0]{{\boldsymbol{\lambda}} }
\newcommand{\muv}{\boldsymbol{\mu}}
\newcommand{\nuv}{{\boldsymbol{\nu}}}
\newcommand{\xiv}{{\boldsymbol{\xi}}}
\newcommand{\omicronv}[0]{{\boldsymbol{\omicron}} }
\newcommand{\piv}{\boldsymbol{\pi}}
\newcommand{\rhov}[0]{{\boldsymbol{\rho}} }
\newcommand{\sigmav}[0]{{\boldsymbol{\sigma}} }
\newcommand{\tauv}[0]{{\boldsymbol{\tau}} }
\newcommand{\upsilonv}[0]{{\boldsymbol{\upsilon}} }
\newcommand{\phiv}{\boldsymbol{\phi}}
\newcommand{\chiv}[0]{{\boldsymbol{\chi}} }
\newcommand{\psiv}{\boldsymbol{\psi}}
\newcommand{\omegav}[0]{{\boldsymbol{\omega}} }

\newcommand{\xin}[1]{{\textcolor{red}{#1}}}

\newcommand{\ts}{^{\top}}
\newcommand{\TV}{{\rm TV}}
\newtheorem{definition}{Definition}
\newtheorem{lemma}{Lemma}
\newtheorem{corollary}{Corollary}
\newtheorem{theorem}{Theorem}

%% file: main.bbl
\begin{thebibliography}{10}

\bibitem{crossnet}
Haitian Zheng, Mengqi Ji, Haoqian Wang, Yebin Liu, and Lu~Fang,
\newblock ``Crossnet: An end-to-end reference-based super resolution network
  using cross-scale warping,''
\newblock in {\em Proceedings of the European conference on computer vision
  (ECCV)}, 2018, pp. 88--104.

\bibitem{xue2019video}
Tianfan Xue, Baian Chen, Jiajun Wu, Donglai Wei, and William~T Freeman,
\newblock ``Video enhancement with task-oriented flow,''
\newblock {\em International Journal of Computer Vision}, vol. 127, no. 8, pp.
  1106--1125, 2019.

\bibitem{wang2018non}
Xiaolong Wang, Ross Girshick, Abhinav Gupta, and Kaiming He,
\newblock ``Non-local neural networks,''
\newblock in {\em Proceedings of the IEEE conference on computer vision and
  pattern recognition}, 2018, pp. 7794--7803.

\bibitem{efenet}
Yaping Zhao, Mengqi Ji, Ruqi Huang, Bin Wang, and Shengjin Wang,
\newblock ``Efenet: Reference-based video super-resolution with enhanced flow
  estimation,''
\newblock in {\em CAAI International Conference on Artificial Intelligence}.
  Springer, 2021, pp. 371--383.

\bibitem{krishnan2009fast}
Dilip Krishnan and Rob Fergus,
\newblock ``Fast image deconvolution using hyper-laplacian priors,''
\newblock {\em Advances in neural information processing systems}, vol. 22, pp.
  1033--1041, 2009.

\bibitem{zoran2011learning}
Daniel Zoran and Yair Weiss,
\newblock ``From learning models of natural image patches to whole image
  restoration,''
\newblock in {\em 2011 International Conference on Computer Vision}. IEEE,
  2011, pp. 479--486.

\bibitem{babacan2008total}
S~Derin Babacan, Rafael Molina, and Aggelos~K Katsaggelos,
\newblock ``Total variation super resolution using a variational approach,''
\newblock in {\em 2008 15th IEEE International Conference on Image Processing}.
  IEEE, 2008, pp. 641--644.

\bibitem{chang2004super}
Hong Chang, Dit-Yan Yeung, and Yimin Xiong,
\newblock ``Super-resolution through neighbor embedding,''
\newblock in {\em Proceedings of the 2004 IEEE Computer Society Conference on
  Computer Vision and Pattern Recognition, 2004. CVPR 2004.} IEEE, 2004,
  vol.~1, pp. I--I.

\bibitem{yang2008image}
Jianchao Yang, John Wright, Thomas Huang, and Yi~Ma,
\newblock ``Image super-resolution as sparse representation of raw image
  patches,''
\newblock in {\em 2008 IEEE conference on computer vision and pattern
  recognition}. IEEE, 2008, pp. 1--8.

\bibitem{salvador2015naive}
Jordi Salvador and Eduardo Perez-Pellitero,
\newblock ``Naive bayes super-resolution forest,''
\newblock in {\em Proceedings of the IEEE International conference on computer
  vision}, 2015, pp. 325--333.

\bibitem{schulter2015fast}
Samuel Schulter, Christian Leistner, and Horst Bischof,
\newblock ``Fast and accurate image upscaling with super-resolution forests,''
\newblock in {\em Proceedings of the IEEE conference on computer vision and
  pattern recognition}, 2015, pp. 3791--3799.

\bibitem{yang2013fast}
Chih-Yuan Yang and Ming-Hsuan Yang,
\newblock ``Fast direct super-resolution by simple functions,''
\newblock in {\em Proceedings of the IEEE international conference on computer
  vision}, 2013, pp. 561--568.

\bibitem{dong2014learning}
Chao Dong, Chen~Change Loy, Kaiming He, and Xiaoou Tang,
\newblock ``Learning a deep convolutional network for image super-resolution,''
\newblock in {\em European conference on computer vision}. Springer, 2014, pp.
  184--199.

\bibitem{kim2016accurate}
Jiwon Kim, Jung~Kwon Lee, and Kyoung~Mu Lee,
\newblock ``Accurate image super-resolution using very deep convolutional
  networks,''
\newblock in {\em Proceedings of the IEEE conference on computer vision and
  pattern recognition}, 2016, pp. 1646--1654.

\bibitem{tao2017detail}
Xin Tao, Hongyun Gao, Renjie Liao, Jue Wang, and Jiaya Jia,
\newblock ``Detail-revealing deep video super-resolution,''
\newblock in {\em Proceedings of the IEEE International Conference on Computer
  Vision}, 2017, pp. 4472--4480.

\bibitem{unet}
Olaf Ronneberger, Philipp Fischer, and Thomas Brox,
\newblock ``U-net: Convolutional networks for biomedical image segmentation,''
\newblock in {\em International Conference on Medical image computing and
  computer-assisted intervention}. Springer, 2015, pp. 234--241.

\bibitem{flownet}
Alexey Dosovitskiy, Philipp Fischer, Eddy Ilg, Philip Hausser, Caner Hazirbas,
  Vladimir Golkov, Patrick Van Der~Smagt, Daniel Cremers, and Thomas Brox,
\newblock ``Flownet: Learning optical flow with convolutional networks,''
\newblock in {\em Proceedings of the IEEE international conference on computer
  vision}, 2015, pp. 2758--2766.

\bibitem{jaderberg2015spatial}
Max Jaderberg, Karen Simonyan, Andrew Zisserman, et~al.,
\newblock ``Spatial transformer networks,''
\newblock {\em Advances in neural information processing systems}, vol. 28, pp.
  2017--2025, 2015.

\bibitem{kingma2014adam}
Diederik~P Kingma and Jimmy Ba,
\newblock ``Adam: A method for stochastic optimization,''
\newblock {\em arXiv preprint arXiv:1412.6980}, 2014.

\end{thebibliography}
